\documentclass{article}
\usepackage[preprint]{colm2026_conference}
\usepackage[T1]{fontenc}
\usepackage[utf8]{inputenc}
\usepackage{microtype}
\usepackage{hyperref}
\usepackage{url}
\usepackage{graphicx}
\usepackage{booktabs}
\usepackage{multirow}
\usepackage{amsmath}
\usepackage{amssymb}
\usepackage{xcolor}
\usepackage{enumitem}
\definecolor{darkblue}{rgb}{0, 0, 0.5}
\hypersetup{
  colorlinks=true,
  citecolor=darkblue,
  linkcolor=darkblue,
  urlcolor=darkblue,
  pdftitle={When Retrieval Metrics Mislead: Measuring Policy Signal in Long-Horizon Tool-Use Agents},
  pdfauthor={Tianyu Ding and Juan Pablo De la Cruz Weinstein}
}

\title{When Retrieval Metrics Mislead:\\
Measuring Policy Signal in Long-Horizon Tool-Use Agents}

\author{Tianyu Ding\\
Amazon Web Services\\
\texttt{tianyd@amazon.com}
\And
Juan Pablo De la Cruz Weinstein\\
Amazon Web Services\\
\texttt{jcruam@amazon.com}}

\begin{document}
\maketitle
\lhead{Accepted at the Lifelong Agents Workshop (LLA) at COLM 2026}

\begin{abstract}
Exact-match retrieval recall is often used as a proxy for whether a retriever supplies useful policy context to a
downstream decision model. We test this proxy for pre-action policy classification in $\tau$-bench using
Qwen2.5-3B/7B classifiers. Under gold-policy conditioning, a compact structured state improves macro-F1 over raw
trajectories by $0.20$ after tuning at 3B, with the same ordering at 7B under shared hyperparameters. We then replace the benchmark-designated governing rule with the
top-ranked benchmark assertion retrieved from decision-time context. Although the exact governing rule is retrieved at rank~1
for only $7\%$ of airline states, the primary 3B classifier obtains macro-F1 $0.58$ with retrieved assertions versus
$0.60$ with the gold rule ($\Delta=-0.02$, task-cluster 95\% CI $[-0.23,+0.21]$); random non-gold and no-assertion controls score
$0.32$ and $0.21$. We do not detect a macro-F1 difference between retrieved assertions and the gold rule in this configuration,
although the interval remains too wide to establish non-inferiority. The same qualitative pattern appears with a second
retriever and at 7B, while varying across fine-tuning configurations. These results show that exact-match recovery of
the benchmark-designated rule can underestimate the downstream utility of retrieved benchmark assertions in this
setting. Retrieval should therefore be evaluated inside the classification loop rather than by exact-match recall alone.
\end{abstract}

\section{Introduction}
Policy-constrained tool-use systems may retrieve a policy clause before deciding whether a proposed action is
permissible, requires more evidence, or should be blocked. Retrieval is commonly evaluated by whether the
benchmark-designated clause appears among the top-ranked results. When exact-match recall is used as a proxy for
downstream utility, it effectively treats nonmatching retrieved assertions as uninformative. We test that assumption in
$\tau$-bench.

We train Qwen2.5-3B/7B classifiers on pre-action states from airline trajectories. The study separates two
interventions: how decision context is represented when the benchmark-designated policy is supplied, and how
classifier performance changes when the supplied assertion is retrieved rather than gold. This design distinguishes
representation effects from the validity of exact-match governing-rule recall as a measure of retrieval utility:
\begin{quote}
\emph{(Q1) Under gold-policy conditioning, how does decision-state \textbf{representation} affect policy
classification?} \quad
\emph{(Q2) When benchmark assertions are retrieved, does exact recovery of the designated rule predict downstream
classifier performance?}
\end{quote}
For Q1 we hold the classifier, training data, and recipe fixed and vary \emph{only} how the decision state is
represented. We find that a compact \textbf{structured decision state}---an explicit elicitation of \{request intent,
evidence read so far, applicable policy assertion, pending action class\}---substantially improves macro-F1 over raw
trajectory text and avoids majority-class prediction. This is a
gold-policy representation result: the state constructor is given the correct applicable policy. For Q2 we replace
that gold policy with a retrieved benchmark assertion at test time. Exact-match retrieval recall is low---an offline
gold-injection diagnostic predicts the structured advantage needs high exact-clause access, yet off-the-shelf
retrievers recover the exact clause only $0.07$ of the time at rank~1. That diagnostic, however, is not the pipeline
evaluation. When we run the direct retrieval intervention, the primary classifier shows no detectable macro-F1
difference vs.\ the benchmark's gold rule and beats both random non-gold and no-assertion controls by wide margins.
\textbf{For this reported configuration, exact-match recall is too pessimistic: nonmatching retrieved benchmark
assertions can preserve decision-relevant context even when they are not alternate governing rules.} The
effect is sensitive to fine-tuning configuration: the gold-policy gap re-opens for under-trained and
maximally-discriminating classifiers. The supported claim is local to the reported configuration rather than universal.

\paragraph{Contribution and scope.} We make three contributions. First, we isolate the effect of decision-state
representation under gold-policy conditioning and show that structured states improve Qwen2.5-3B/7B macro-F1 over raw
trajectories. Second, we characterize exact-match governing-rule retrieval across two $\tau$-bench domains, multiple query
constructions, and four retrievers, and use a gold-injection probe to quantify the prediction implied by treating
nonmatching assertions as uninformative. Third, we test that prediction directly with gold, retrieved, random non-gold, and
absent assertion inputs. The direct intervention shows that low exact-match recall can coexist
with a small observed gold--retrieved performance gap in the primary configuration, while the effect varies with fine-tuning.
Together, these analyses evaluate the criterion validity of exact-match recall for policy-conditioned decision
models. All claims concern benchmark-authored $\tau$-bench assertions and offline classification.

\paragraph{Relation to long-lived agents.} Here, ``long-horizon'' describes the length and tool-interaction structure
of the source $\tau$-bench trajectories; we do not study continual learning or memory across sessions. The connection
to long-lived agents is through reliability and evaluation: such agents may depend on policy-conditioned gates, and
exact-match recall can misstate the utility of the context supplied to those gates.

\section{Setup}
\paragraph{Task and data.} We use pre-action decision states extracted from $\tau$-bench-airline
\citep{yao2024taubench} trajectories. Each state has a gold decision $a^\star\in\{\texttt{allow},\texttt{verify},
\texttt{refuse}\}$ derived from the task's benchmark assertions and the evidence visible at the checkpoint. We use a
fixed split grouped by \texttt{task\_id} (train $195$ / test $85$; $15$ disjoint test tasks; zero task overlap), so
no test task's assertion strings appear in training.

\paragraph{Construct scope.} The candidate pool contains the benchmark-authored natural-language assertions attached
to each $\tau$-bench task, including governing rules and task-specific state or booking assertions. We therefore use
``benchmark assertion'' for a retrieved candidate and ``governing rule'' for the benchmark-designated assertion that
determines the decision. The identification task is a benchmark proxy: recover that designated rule from
decision-time context over the domain's assertion pool. This tests whether the classifier's gold-rule input is
obtainable inside the benchmark, not whether real policy-document retrieval would be equally easy or hard.

\paragraph{Representations (the independent variable).} Holding everything else fixed we vary the classifier input:
\textbf{raw} (full conversation + raw tool outputs, $\sim$2118 tokens); \textbf{masked} (tool outputs replaced by
typed placeholders, $\sim$717); \textbf{structured} (explicit \{request, evidence, policy, action-class\} fact
extraction, $\sim$185); and \textbf{raw+policy} (raw with the elicited policy assertion appended). The structured
state is produced by a generator that reads the trajectory and the task policy spec; \textbf{eliciting the applicable
policy assertion is an explicit part of the method's cost---the policy input is supplied by the state-construction
procedure rather than inferred by the classifier itself.}

\paragraph{How the structured state is built (and what it does \emph{not} use).} The generator sees only information
available at the checkpoint: the dialogue so far, parsed tool outputs read up to that point, the pending action's
type (read vs.\ write/transfer), and the task's natural-language assertions. \emph{request} is the user's
stated goal; \emph{evidence} lists parsed facts from tool results (e.g.\ \texttt{cabin: basic\_economy; insurance:
no}); \emph{policy} is the applicable assertion lifted verbatim from the task spec (e.g.\ ``Agent should not offer
compensation unless the user asks''); \emph{action class} is read vs.\ write. Crucially the generator does
\textbf{not} see the gold decision $a^\star$, and the policy assertion is a \emph{rule}, not the answer---it
keyword-matches the gold action only 14--33\% of the time (Section~\ref{sec:controls}). This is
\emph{decision-relevant} state, not \emph{label-derived} state.

\paragraph{Decision models and evaluation.} The primary model is a supervised fine-tuned Qwen2.5-3B/7B classifier
\citep{qwen2024qwen25} with LoRA \citep{hu2022lora}
($r{=}16$, $5$ epochs, lr $5\!\times\!10^{-5}$, bf16), which predicts whether the proposed action may proceed,
requires additional verification, or should be blocked. This is the same fine-tuning recipe under which the documented
majority-class failure occurred. For computationally inexpensive diagnostic analyses, we also fit a balanced
logistic-regression classifier over frozen MiniLM embeddings \citep{reimers2019sbert}. The metric is 3-way
macro-F1 (the frozen-encoder structured$>$raw gap given gold policy holds for MiniLM and \texttt{bge-large} but not
\texttt{e5-large}; the Qwen classifier is our primary representation evidence---Appendix~\ref{app:robust}). We
operationalize downstream retrieval utility as macro-F1 when a retrieved benchmark assertion replaces the
benchmark-designated governing rule in the classifier input.
Confidence intervals are nonparametric bootstraps. Representation-table intervals resample the per-example test set.
For the direct retrieval intervention, point estimates average per-seed macro-F1 values; paired-delta CIs are
task-cluster bootstraps over the $15$ test task IDs ($5000$ resamples), recomputing and averaging seed-level deltas
per replicate. SFT results pool the $3$
seeds' test predictions in the representation tables. We call a classifier
\emph{collapsed} when it assigns one class to $\ge 90\%$ of test items (the documented
\texttt{ce\_smoke} failure is \texttt{refuse}$\approx$100\%).

\section{Policy-conditioned representation ablation}
\label{sec:results}

Holding the classifier, training data, and recipe fixed, we vary \emph{only} the input representation. In this section
the classifier is given the correct applicable policy clause; we replace this gold-policy condition in
\S\ref{sec:identification}. The main result is that a compact \textbf{structured decision state}
($\sim$185 tokens) that surfaces the applicable policy reaches macro-F1 $0.603$ at 3B vs.\ $0.283$ for raw trajectory
text ($\sim$2118 tokens)---a paired gap of $+0.320$ (CI $[+0.244,+0.397]$), which is not an end-to-end gain but a
representation effect \emph{given} the rule. The advantage remains $+0.199$ after 3B tuning on a held-out validation
split and is $+0.157$ at 7B under shared hyperparameters (Table~\ref{tab:scale} in
Appendix~\ref{app:repr}).

The effect is not brevity (length/info-matched controls), not a leaked label (a mismatched policy sharply reduces
performance), and not output-format compliance (it survives parsed-only rescoring). Among the tested representation
interventions, explicit policy access produces the largest observed improvement: appending the applicable policy to raw text helps nearly
as much as the full structured state, and a mismatched policy is worse than no policy. Cross-actor transfer
(Sonnet$\to$Nova) is positive but not
significant at $n{=}50$ (Appendix~\ref{app:repr}).

\textbf{Critical scope:} these results are gold-policy conditioned: the structured state contains the correct
benchmark policy clause. Section~\ref{sec:identification} asks whether the classifier still works when that clause is
\emph{retrieved} rather than supplied by the benchmark.

\section{Evaluating exact-match recall as a proxy for downstream classifier utility}
\label{sec:identification}
Every result so far conditions on the classifier receiving the applicable benchmark policy assertion. The downstream
measurement question is whether exact-match retrieval recall predicts classifier performance when that assertion is
\emph{retrieved} from decision-time context. The concern is that retrieval may be too weak: off-the-shelf retrievers
recover the exact governing clause only rarely (recall@1 $\approx0.07$). We first run an exact-recall diagnostic, then
\emph{feed retrieved assertions directly into the classifier} rather than inferring downstream utility from recall alone.
We then map broader exact-match recall patterns across domains, protocols, and retrievers. The reported Qwen
classifier is more robust to imperfect retrieval than exact-recall diagnostics alone would suggest.

\begin{figure}[t]
\centering
\includegraphics[width=\linewidth]{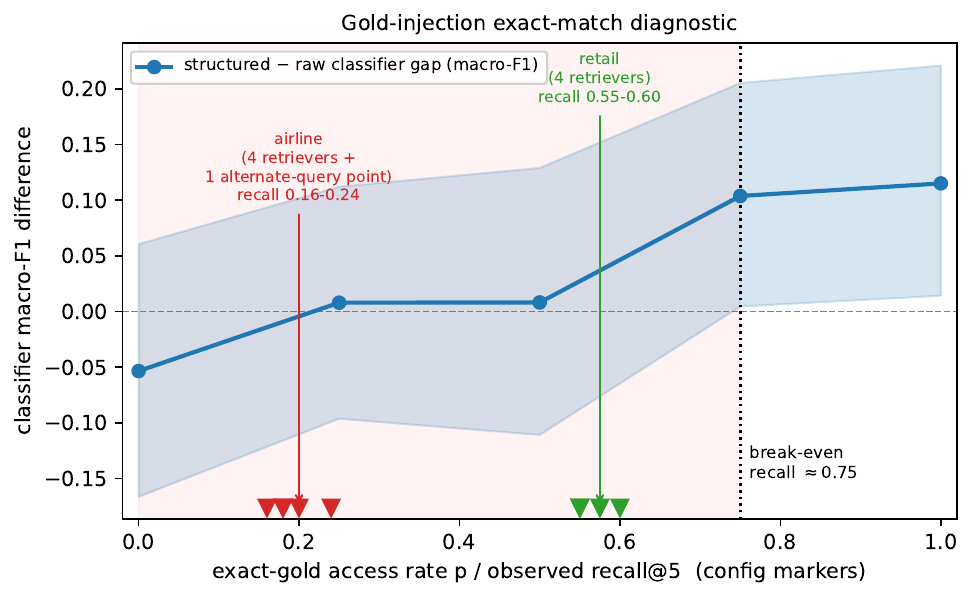}
\caption{\textbf{Gold-injection exact-match diagnostic.} The curve
is the structured$-$raw classifier macro-F1 gap as a function of effective \emph{exact-gold access rate}
(offline gold-injection sweep: gold clause for a fraction $p$ of states, MiniLM top-1 retrieved assertion otherwise; $95\%$ CI
band); it crosses zero only near recall $\approx0.75$. Triangles mark \emph{achievable} exact-match recall@5 for four
retrievers under trajectory-context in each domain, plus one airline request+evidence MiniLM point. All lie far left of the crossing, which
\emph{predicts} substantially lower downstream performance if nonmatching retrieved assertions are treated as
low-utility. Table~\ref{tab:retrievedgate} tests that prediction directly by replacing the gold governing-rule input
with retrieved assertion inputs.}
\label{fig:breakeven}
\end{figure}

\paragraph{Gold-policy injection diagnostic.} We hold the diagnostic probe, data, and recipe fixed and degrade only the policy
input: for a fraction $p$ of states the classifier receives the gold clause, otherwise it receives the MiniLM top-1 retrieved
clause from the same retrieval pool. This gold-injection sweep makes $p$ an \emph{effective exact-gold access rate}.
We run this on the frozen-encoder diagnostic probe (whose gold-policy structured$-$raw gap is $+0.115$). On a coarse
grid the gap first becomes robustly
positive near $p\!\approx\!0.75$ ($+0.104$, $95\%$ CI $[+0.004,+0.205]$); on a finer grid (steps of $0.1$, three
injection seeds) the gap's CI excludes zero only at $p\!=\!1.0$ (with $p\!=\!0.8$ marginal, CI lower $-0.001$), i.e.\
the finer sweep places the threshold \emph{higher}, not lower (and it is robust to the classifier head---a more
regularised logistic head and a small MLP never clear zero even at full injection; Appendix~\ref{app:robust}).
\textbf{Either way, this diagnostic probe says the structured-vs.-raw advantage becomes robust only when exact gold-clause
access is high---at least $\sim$$0.75$} (Figure~\ref{fig:breakeven}). We treat this as an \emph{order-of-magnitude}
diagnostic, not a sharp pipeline threshold: it is estimated on a frozen-encoder probe, the finer grid puts it if
anything \emph{higher}, and the Qwen classifier's larger gold-policy representation gap could behave
differently. The direct intervention below asks whether this exact-recall diagnostic predicts the primary Qwen
configuration.

\paragraph{The proxy assumption.} The diagnostic's $p$ is an exact-gold access rate inside a frozen-encoder probe; the
retrieval number is exact-match recall@$k$; the direct intervention is the Qwen classifier with retrieved assertions in
the input. Treating the diagnostic as a deployment prediction assumes that exact governing-rule recovery is a close proxy
for downstream classifier utility. The direct intervention tests that assumption and finds it too pessimistic for the
reported classifier: a non-gold but decision-context-aligned retrieved assertion can be useful, so exact-match recall is a
poor stand-alone proxy for downstream classifier utility in this configuration. The diagnostic is pessimistic not
because its non-gold inputs are random---they are MiniLM top-1 retrieved assertions---but because the probe treats those
non-gold retrieved assertions as low-utility, while the direct classifier intervention measures their downstream effect.

We establish the diagnostic crossing point with a \emph{controlled} gold-injection sweep (varying identification while holding
everything else fixed) rather than an observational ``does retrieval success predict classifier accuracy'' correlation; the
latter is tiny ($11/85$) and confounded with per-example difficulty, and we report it only in Appendix~\ref{app:robust}.

\paragraph{Retrieved-assertion classifier intervention.} The diagnostic crossing point and the recall measurement are
both \emph{indirect}: they infer a downstream outcome from how often retrieval
returns the \emph{exact} gold rule. Table~\ref{tab:retrievedgate} reports the central direct test on the
retrieval-conditioned Qwen classifier. We train on the benchmark's gold structured states, then at test time replace
the assertion line with the \emph{top-1 retrieved} benchmark assertion (recall@1 $0.07$), the gold governing rule, a
random non-gold assertion, or no assertion line at all. On the primary
configuration (Qwen2.5-3B, the same configuration that reaches structured macro-F1 $0.60$ given the gold rule), the
retrieved assertion yields macro-F1 $0.580$ versus $0.604$ for the benchmark's gold rule---a gap of $-0.024$
($95\%$ CI $[-0.233,+0.207]$): \emph{no detectable degradation in this experiment} (we prespecify a
non-inferiority margin $\delta{=}0.05$ macro-F1, $\sim$$10\%$ of the gold$-$raw gap; at $n{=}85$ the gap's CI is too
wide to establish non-inferiority, so we report ``no detectable loss,'' not equivalence). The retrieved assertion
sits far above both a random non-gold assertion ($0.315$; top-1 retrieved$-$random $+0.265$, CI $[+0.056,+0.440]$) and a
classifier given \emph{no} assertion line ($0.206$, which falls into a majority-\texttt{verify} regime); top-1 over
no-assertion is $+0.374$ (CI $[+0.121,+0.569]$, excludes zero). \textbf{Thus the retrieved assertion supplies measurably
useful decision context in this classifier evaluation despite recovering the exact gold string only $7\%$ of the time.}

\begin{table}[t]
\centering
\small
\begin{tabular}{lccc}
\toprule
Assertion input at test time & Macro-F1 & $\Delta$ vs.\ gold & 95\% CI \\
\midrule
Gold governing rule & 0.604 & -- & -- \\
Top-1 retrieved (MiniLM, recall@1 $0.07$) & 0.580 & $-0.024$ & $[-0.233,+0.207]$ \\
Top-1 retrieved (\texttt{bge-large}, recall@1 $0.07$) & 0.571 & $-0.033$ & $[-0.338,+0.269]$ \\
Random non-gold assertion & 0.315 & $-0.288$ & $[-0.543,-0.017]$ \\
No assertion line & 0.206 & $-0.398$ & $[-0.647,-0.096]$ \\
\bottomrule
\end{tabular}
\caption{\textbf{Effect of assertion input on the primary Qwen2.5-3B classifier.} The classifier is trained on gold structured states;
at test time only the assertion line changes. Macro-F1 is the mean of per-seed macro-F1 values over three seeds on the
same $n{=}85$ airline states. Delta intervals are task-cluster paired bootstraps over the mean seed-level delta.
The top-1 retrieved assertion shows no detectable loss vs.\ the gold governing rule, although the CI is compatible with
a meaningful degradation. Random non-gold and no-assertion controls show that the retrieved assertion contributes
decision-relevant context.}
\label{tab:retrievedgate}
\end{table}

\paragraph{Analysis of nonmatching retrieved assertions.} The mechanism evidence points to context alignment rather
than recovery of an alternate governing rule. The top-1 retrieved assertion is no closer to the gold clause
in embedding space than a random assertion is (cosine $0.39$ vs.\ $0.40$), but it is far more similar to the request$+$evidence context than a
random assertion ($0.55$ vs.\ $0.26$), and the assertion pool is not merely redundant (mean pairwise cosine $0.39$,
$72/122$ clusters at cosine $\ge0.8$). A single-annotator assertion categorization
(Appendix~\ref{app:clausecoding}) labels $59$ of the $79$ nonmatching states as \emph{inapplicable} rules and the
remaining $20$ as \emph{sufficient} or \emph{partially relevant}. Those $20$ states represent five test tasks and five
unique gold/retrieved assertion pairs; the $10$ sufficient states come from two tasks and two unique pairs. The no-loss
result therefore appears to combine tolerance of context-aligned non-rule text with a small number of useful alternate
rules. An additional control tempers this interpretation: the retrieved assertions are also
somewhat longer than the random non-gold ones ($21$ vs.\ $10$ tokens), so we cannot fully separate context alignment from a
length effect; we report this as a partial confound rather than a clean mechanism. Clause length alone cannot explain
the assertion-input effects: the gold and random non-gold assertions are essentially equal length (both $\sim$$10$ tokens) yet
differ by $+0.29$ macro-F1. However, length may still contribute to the retrieved-versus-mismatched contrast, so the
context-alignment analysis should be interpreted as suggestive rather than causal. A length-matched
retrieved-versus-random control (padding or truncating assertions to equal length) would isolate the length
contribution and is left to future work. The effect is also
\textbf{sensitive to fine-tuning configuration}: the gold-policy gap stays near zero across the mid-range classifiers that reach $\sim$$0.60$ given
gold (including a second, different-family retriever, \texttt{bge-large}, and at 7B), but it \emph{re-opens} for
under-trained classifiers (gold $\le0.49$; top-1$-$gold down to $-0.16$) and for the single most aggressively-trained classifier
we tried (8 epochs, lr $10^{-4}$; gold $0.63$, top-1$-$gold $-0.15$). The result is therefore scoped to the reported
configuration: in this regime, low exact-match recall does not imply low utility for retrieved benchmark assertions. (The low-cost
frozen-encoder diagnostic probe shows the same pattern; seed-level 3B/7B direct-retrieval results appear in
Appendix Table~\ref{tab:directpolicyseeds}.)

\paragraph{Exact-match retrieval performance.} We then measure how often the applicable clause is exactly
retrieved, framed as
retrieval over the domain's full natural-language assertion set (the ``haystack''): for each decision we rank
all assertions by similarity to the available context and record whether the gold governing rule is in the top-$k$.
We do this for \textbf{two public domains} ($\tau$-bench airline, $122$ assertions; retail, $51$ assertions), \textbf{two
query-construction protocols} (the request$+$evidence decision state; and pooled pre-action user context
from the trajectory), and \textbf{four off-the-shelf retrievers}: a MiniLM bi-encoder, two stronger instruction-tuned
bi-encoders (\texttt{bge-large-en-v1.5}, \texttt{e5-large-v2}), and a cross-encoder reranker
(\texttt{bge-reranker-large}) scoring the \emph{entire} assertion pool (the ceiling of a first-stage$+$rerank pipeline).
\textbf{All evaluated configurations remain below the diagnostic crossing}: across the four retrievers, recall@5 is $0.18$--$0.24$ on
airline and $0.55$--$0.60$ on retail (and $0.16$ on airline under the leaner request$+$evidence query protocol); the
strongest retriever (the cross-encoder reranker) does not change the picture ($0.18$ airline, $0.57$ retail). The
shortfall persists even at larger $k$: the best recall@10 across retrievers is only $0.36$ (airline) and $0.72$
(retail), and the full recall@$k$ curves (Appendix Figure~\ref{fig:recallcurves}) stay below the diagnostic crossing until
$k$ reaches large candidate sets comprising tens of clauses. The
gold clause has a poor median rank of approximately $45$ out of $122$ on airline. The identification shortfall is therefore
\emph{not} an artifact of a weak bi-encoder---it is robust to retriever strength. We also go beyond off-the-shelf
retrieval and test three additional retrieval pipelines---query expansion, a hybrid first stage with
cross-encoder reranking of the top candidates, and a coarse-to-fine hierarchical retrieve---none of which clears the
diagnostic crossing either (best recall@5 $0.38$ airline / $0.58$ retail; the hierarchical variant is actually worse, as
coarse section routing drops the gold clause; Appendix~\ref{app:robust}). So the exact-match shortfall is not merely
a weak-retriever artifact: it survives stronger, structured identification. Retail is the closest call: at
recall@5 $\approx0.55$--$0.60$ it sits closest to the threshold of any configuration, though still below it (and the
higher finer-grid threshold places retail more clearly below); we nonetheless flag retail as the near-miss
case in exact-recall terms rather than a decisive failure.

Exact-match recall also varies substantially by decision class (Appendix~\ref{app:robust}, Table~\ref{tab:identifclass}):
recall@5 is $0.00$ for \texttt{allow}, $0.05$ for \texttt{verify}, and $0.50$ for \texttt{refuse}. Even for
\texttt{refuse}, the benchmark-designated clause is absent from the top five in approximately half of the evaluated
states.

\paragraph{Sensitivity to author-side scenario fields.} A third query construction---appending the task's
\texttt{task\_instructions}/\texttt{known\_info} scenario fields---reaches higher recall (up to $0.58$ airline,
$0.78$ retail, the latter nominally clearing the diagnostic crossing). We \emph{exclude} it because those fields are written from
the task author's knowledge of the correct resolution and leak policy-relevant content into the user-side query, and
because our scoring there credits the best of multiple gold clauses (an upper bound). It does not reflect what a classifier
can see at decision time. We report it for completeness; the two main protocols
use only decision-time-available context. The two main sweeps differ in construction---the request$+$evidence figure is $n{=}85$ airline
states, the four-retriever robustness sweep uses the trajectory-context protocol at $n{=}50$---but both, and retail,
sit below the diagnostic crossing.

\paragraph{Sensitivity to domain and query construction.} Airline ($\approx0.16$) looks much harder than retail
($\approx0.55$), which would suggest identification difficulty is \emph{domain-dependent}---and indeed, when the two
domains are queried with \emph{different} protocols, their gold-rank-percentile confidence intervals separate. But
that separation disappears under matched query construction, suggesting sensitivity to the query protocol rather than
evidence for a stable domain difference. In a controlled $2{\times}2$
(domain $\times$ query protocol), applying the \emph{same} trajectory-context protocol to both domains does not yield a
detectable difference (gold-rank percentile $0.10$ airline vs.\ $0.09$ retail; Mann--Whitney $p{=}0.29$,
$n{=}50/40$). This is a \emph{failure to detect} a difference at this sample size, not proof of equivalence, so the
defensible statement is that the domains are \emph{not distinguishable under a matched
protocol}, and the large cross-domain gap one might report appears only when the two domains are queried differently.
The supported statement is one of \emph{consistency} (not domain-invariance): achievable
exact-match recall stays below the offline gold-injection diagnostic for all four retrievers under the trajectory-context
protocol in both domains, as does an additional airline request+evidence point (Figure~\ref{fig:breakeven}).

\paragraph{Interpretation.} The representation effect (Section~\ref{sec:results}, full ablation in
Appendix~\ref{app:repr}) is conditioned on access to the benchmark-designated policy clause. The remaining question is
whether retrieved assertion inputs preserve this representation advantage---and Steps~1 and~3 confirm that exact-match
recall \emph{is} low. But the direct test
(Step~2) shows that the recall shortfall does not translate into a detectable loss on the reported classifier:
\textbf{for this classifier, the top-1 retrieved assertion shows no detectable macro-F1 loss vs.\ the benchmark's gold
clause despite low exact-match recall}, so exact-match recall@k would substantially understate the utility of retrieved assertions in
this setting if used alone. Two bounds matter:
the effect is sensitive to fine-tuning configuration (the gold-policy gap re-opens for under-trained and maximally-discriminating classifiers,
so low exact-match recall can still translate into detectable losses in those regimes), and at $n{=}85$ we report ``no detectable degradation'' rather
than equivalence or established non-inferiority. This reframes the evaluation target: report retrieved-assertion
classifier performance alongside exact-rule recall, and reserve high-stakes guarantees for systems with verification
beyond retrieval alone. Appendix Table~\ref{tab:provenance} maps each main identification number to its result file.

\section{Related work}
\paragraph{Tool-use agents and intermediate state.} Our classifier targets the pre-action checkpoint of the
reason-then-act loop popularized by ReAct \citep{yao2023react}, where an agent interleaves reasoning and tool calls.
The idea of decisions riding on an explicit \emph{structured intermediate state} has deep roots in task-oriented
dialogue, where belief/dialogue-state tracking maintains a structured (and ideally calibrated) state distribution
that downstream policy decisions consume \citep{vanniekerk2020belief}; SSDG's compliance-oriented decision state is
a close analogue for write-action gating. Deciding \emph{whether to act at all}---to abstain, refuse, or ask---is
itself studied as a first-class capability: abstention abilities of LLMs \citep{madhusudhan2024abstention} and
asking clarifying questions before acting \citep{wu2023clarifying} both motivate our three-way
allow/verify-then-proceed/refuse output rather than a binary block/allow classifier.

\paragraph{Compliance gating.} Pre-action compliance gating for tool agents is dominated by two non-learned
paradigms: symbolic/formal runtime
enforcement that compiles policies into rules or SMT constraints and intercepts tool calls
\citep{shieldagent2025,solveraided2026}, and prompt/document-level policy conditioning, the $\tau$-bench default.
Closest to our structured-state setting, \citet{zwerdling2025companypolicy} compile company-policy documents into per-tool
guard code on $\tau$-bench-airline; like the symbolic enforcers, this presupposes the applicable policy is already
in hand and does not ask whether the governing clause can be \emph{identified} at decision time---the question we
isolate here.
Retrieval-based guards \citep{xiang2024guardagent} and agentic-RAG policy-as-code \citep{romeo2025arpaccino} do
retrieve guard knowledge, and rule-based runtime-enforcement DSLs \citep{wang2025agentspec} sit in our pre-action
allow/refuse design space, but all presuppose the rule and none measures identification recall against downstream
classifier utility. The nearest compliance-benchmark neighbor, MANTRA \citep{anand2026mantra}, synthesizes SMT-validated
compliance traces but verifies behavior \emph{post hoc} rather than gating actions pre-emptively.
Learned guards also classify over \emph{raw} trajectory text against harm taxonomies
\citep{inan2023llamaguard} or raw prefixes. The closest representation-axis work, sufficient-context for RAG
\citep{joren2024sufficient}, distills a single sufficiency signal for generate-vs-abstain, not a multi-field policy
state for write-action gating. Recent agent-safety work motivates but does not run our ablation: the
``verifier tax'' shows runtime enforcement intercepts most unsafe actions yet rarely yields safe completion
\citep{sah2026verifiertax}, and proxy-state evaluation infers a structured state \emph{post hoc} for reward/eval
rather than as a pre-action classifier input \citep{chuang2026proxystate}. To our knowledge, among the venues surveyed
above, we are the first to run a controlled \emph{decision-state representation ablation} (model$+$data$+$recipe
fixed) specifically for a \emph{learned allow/verify/refuse pre-action classifier in policy-constrained tool-use agents}
(as opposed to guardrails or abstention in general).

\paragraph{Exact-match recall vs.\ downstream policy utility.} A prominent line treats an upstream retrieval/selection step, not
the downstream model, as what bounds end-to-end quality: for \emph{tools}, \citet{shi2025toolret} show strong
retrievers select the right tool poorly and that benchmarks mask this by pre-annotating the relevant tool; for
\emph{skills}, \citet{wang2026notallskills} find the binding constraint on $\tau$-bench is matching skills to tasks;
and task-aligned retrieval \citep{sun2026taskaligned} argues one should retrieve the \emph{applicable} item, not the
most lexically similar one. Our direct intervention partially \emph{contrasts} with this line: at the level of
\emph{exact} clause recall our retriever is just as weak ($7\%$ at rank~1), yet the reported classifier does not show
a detectable macro-F1 loss because a non-gold but decision-context-aligned clause can remain useful. Prior work has
established that retrieval relevance and exact-match metrics can be weak proxies for downstream utility
(sufficient-context for RAG \citep{joren2024sufficient}; answer-equivalence shows token-level exact-match
\emph{underestimates} correct \emph{outputs} \citep{bulian2022tomayto}; and recall is itself a problematic
retrieval-quality metric whose link to LLM response quality is weak \citep{schwartz2025recall}). We evaluate this
decoupling in a new setting: retrieval of benchmark-authored policy clauses for pre-action compliance classification.
The compliance-enforcement literature largely
\emph{assumes} the rule is in hand---symbolic enforcers compile a \emph{known} policy \citep{shieldagent2025,
solveraided2026}, proxy-state evaluators infer state for a \emph{given} task \citep{chuang2026proxystate}---and where
it does retrieve, it (correctly) warns that retrieval alone gives no high-stakes guarantee; our claim is correspondingly
about the reported classifier's \emph{accuracy}, not a safety guarantee. Closest in spirit to our
mechanism, \citet{chen2026lookingnotpicking} argue tool-selection failure is at the decision readout, not
identification. In our setting, the corresponding observation is that the classifier can exploit a context-aligned
nonmatching clause. To our knowledge, the
specific measurement---that exact-match recovery of the designated rule undercounts the downstream utility of
retrieved benchmark assertions for a compliance classifier, and that this varies with fine-tuning configuration---has
not been reported.

\section{Limitations}
\label{sec:limits}
\textbf{Construct validity: a benchmark-native proxy for ``the rule,'' not real policy documents.} Our retrieval
corpus is the set of each domain's distinct $\tau$-bench \texttt{nl\_assertion} strings ($122$ airline, $51$ retail).
It includes governing rules as well as task-specific state and booking assertions; the gold labels and gold-rule input
are drawn from the benchmark-designated assertions. This is a \emph{proxy} for rule identification: we measure whether
the designated governing \emph{assertion} can be recovered from
decision-time context, not whether an agent can identify the applicable clause from a naturalistic, multi-page policy
document. This scope has two implications: (i) external validity to real policy corpora is \emph{unproven}---a longer,
redundant, or differently-structured policy document could be easier or harder; and (ii) our stronger-identifier
results (query expansion, cross-encoder reranking, hierarchical retrieval) are robustness checks \emph{within this same
surrogate construct}, not evidence about realistic policy-document identification. We therefore frame the contribution
as a benchmark-scoped metric-validity study: exact-match recall can be too pessimistic a proxy for downstream
classifier utility, rather than a general law about tool-use agents or policy documents.

\textbf{Exact recovery is not equivalent to downstream utility.} We show exact-match governing-rule recall falls below the
offline gold-injection diagnostic across the configurations we test (four off-the-shelf retrievers under trajectory-context
in two domains, plus one airline request+evidence point), while the direct classifier evaluation shows that low exact recall need not imply low downstream
utility. The shortfall is scoped to \emph{training-free, off-the-shelf} retrieval; a domain-fine-tuned dense retriever,
a learned clause selector, or an agent-in-the-loop that asks clarifying questions could improve exact recovery. More
importantly, the direct intervention shows that exact recovery is not the only utility metric: retrieved-assertion-in-the-loop
classifier accuracy must be measured directly. The diagnostic crossing ($\approx0.75$ on a coarse grid; the finer grid puts
it higher) is therefore an order-of-magnitude warning about exact recall, not a sharp deployment threshold.

\textbf{Scale and scope.} This is a small-$N$ study of a pre-action classifier evaluated offline rather than inside a live
agent loop. The representation half is on $\tau$-bench-airline (test $85$, $15$ tasks; cross-actor $n{=}50$); the
identification half adds retail ($n{=}40$ tasks carrying policy assertions). Two domains is the ceiling of the
public benchmark substrate, not a convenience sample: of the available $\tau^2$-bench domains, only airline ($50/50$
tasks) and retail ($40/114$) expose the per-task natural-language policy assertions our labels and haystack require;
telecom ($0/2285$) and banking ($0/97$) expose none, so a third domain would require generating new trajectories
(reintroducing the actor confound) and is out of scope for a training-free study. The retail trajectories come from
different actor models
than the airline decision states, so we keep the cross-domain claim at the actor-agnostic
\emph{identification} level (retrieval recall and rank), not at absolute classifier macro-F1. We report two model scales
(3B, 7B) under shared hyperparameters and a per-cell tuned 3B sensitivity analysis on a small held-out split ($55$
states / $10$ tasks), reporting ``best observed configuration'' rather than a tuned optimum. The 7B ranking of
\emph{masked} vs.\ \emph{structured} under shared hyperparameters should not be over-read. The evidence supports a
bounded claim: given the benchmark policy, raw trajectory text is a poor classifier input, and in the primary
configuration a top-1 \emph{retrieved} assertion showed
no detectable degradation vs.\ the benchmark gold clause despite low exact-match recall. Retrieval is not universally
sufficient: the result is sensitive to fine-tuning configuration (the gold-policy gap re-opens for under-trained and
maximally-discriminating classifiers), the test sets are small with benchmark-derived labels, and the
decision-context-alignment mechanism is partly confounded with retrieved-clause length and label coarseness
(Appendix~\ref{app:clausecoding}). Live-loop deployment, learned identification, and high-stakes verification
beyond retrieval are left to future work.

\textbf{Upstream extractor cost and error propagation.} A strong upstream LLM builds each structured decision state
by reading the trajectory and eliciting the applicable policy assertion. We do not measure this extraction cost or
how extraction errors propagate to the classifier. A cheaper or less accurate extractor may reduce the observed
representation gain.

\section{Conclusion}
Exact-match governing-rule recall is not a reliable stand-alone measure of retrieval utility for the classifier studied
here. Under gold-policy conditioning, structured decision states outperform raw trajectories, confirming that explicit
policy context is useful. However, replacing the gold rule with a top-ranked retrieved assertion yields similar
macro-F1 in the primary Qwen2.5-3B configuration despite $7\%$ exact recall@1, whereas random non-gold and no-assertion
inputs perform substantially worse. This result is sensitive to fine-tuning and does not establish non-inferiority,
but it shows that nonmatching retrieved assertions can preserve decision-relevant context even when they are not
equivalent policy rules. Evaluations of benchmark rule retrieval for tool-use compliance classifiers should therefore
report downstream performance with retrieved assertions alongside exact-match recall.

\bibliographystyle{colm2026_conference}
\bibliography{references}

\appendix

\section*{Data and reproducibility}
All experiments use public benchmarks: $\tau$-bench \citep{yao2024taubench} and the MIT-licensed $\tau^2$-bench
release \citep{barres2025tau2bench} (airline and retail domains). All trajectories are generated by
publicly-available models on these open-source benchmarks. For the representation half, the airline pre-action
decision states are extracted from $\tau$-bench-airline trajectories produced by Claude-Sonnet-4 and Amazon-Nova-Lite
agents; the classifier that is trained/evaluated on them is a separate small model (Qwen2.5-3B/7B, or a frozen MiniLM
probe). For the identification half, we use the public $\tau^2$-bench rollouts (Claude-3.7/GPT-4.1/o4-mini actors)
for airline and retail. In all cases the gold allow/verify/refuse labels and the policy clauses derive from each
task's published natural-language policy assertions. No customer, production, or proprietary data is used and no
internal models are involved---$\tau$-bench personas are synthetic and all actor/classifier models are public. The
accompanying artifact contains the model outputs and analysis scripts required to reproduce all reported statistics,
including the gold-injection diagnostic sweep and the cross-domain identification measurements (including the excluded
leaky-query protocol, flagged as such).

\section{Identification-number provenance}
\label{app:provenance}
Table~\ref{tab:provenance} maps the main identification numbers to the result files used to compute them.

\begin{table}[t]
\centering\footnotesize
\setlength{\tabcolsep}{3pt}
\begin{tabular}{@{}p{0.44\columnwidth} l l@{}}
\toprule
Headline quantity & Value & Source artifact \\
\midrule
Break-even recall (coarse) & $\approx0.75$ & \texttt{kill\_test\_v2} \\
Break-even (fine grid) & $p{=}1.0$ & \texttt{breakeven\_fine} \\
Oracle struct$-$raw gap (offline) & $+0.115$ & \texttt{kill\_test\_v2} \\
Struct$-$raw (3B tuned) & $+.199$ & \texttt{hp\_search/preds} \\
Struct$-$raw (7B shared) & $+.157$ & \texttt{e6\_7b/preds} \\
Airline R@5, req$+$ev (hybrid) & $0.16$ & \texttt{ret\_ceiling} \\
Airline R@5, traj-ctx ($\times4$) & $0.18$--$0.24$ & \texttt{stronger\_retr} \\
Retail R@5, traj-ctx ($\times4$) & $0.55$--$0.60$ & \texttt{stronger\_retr} \\
Best R@10 (air/ret) & $0.36$/$0.72$ & \texttt{recall\_curves} \\
Matched pctile (air/ret) & $0.10$/$0.09$ & \texttt{protocol\_sym} \\
Matched Mann--Whitney & $p{=}0.29$ & \texttt{protocol\_sym} \\
Per-class R@5 (a/v/r) & $.00$/$.05$/$.50$ & \texttt{identif\_class} \\
Direct MiniLM top-1 vs.\ gold & $-.024$ $[-.233,.207]$ & \texttt{direct\_policy\_stats} \\
Direct top-1 vs.\ no assertion & $+.374$ $[.121,.569]$ & \texttt{direct\_policy\_stats} \\
\bottomrule
\end{tabular}
\caption{Provenance map for the headline numbers: each quantity, its value, and the committed
result artifact (under \texttt{experiments/}) it is recomputed from. The two airline R@5 figures use different query
protocols (request$+$evidence vs.\ trajectory-context), reported separately; all land below the diagnostic crossing.}
\label{tab:provenance}
\end{table}

\begin{figure*}[t]
\centering
\includegraphics[width=\textwidth]{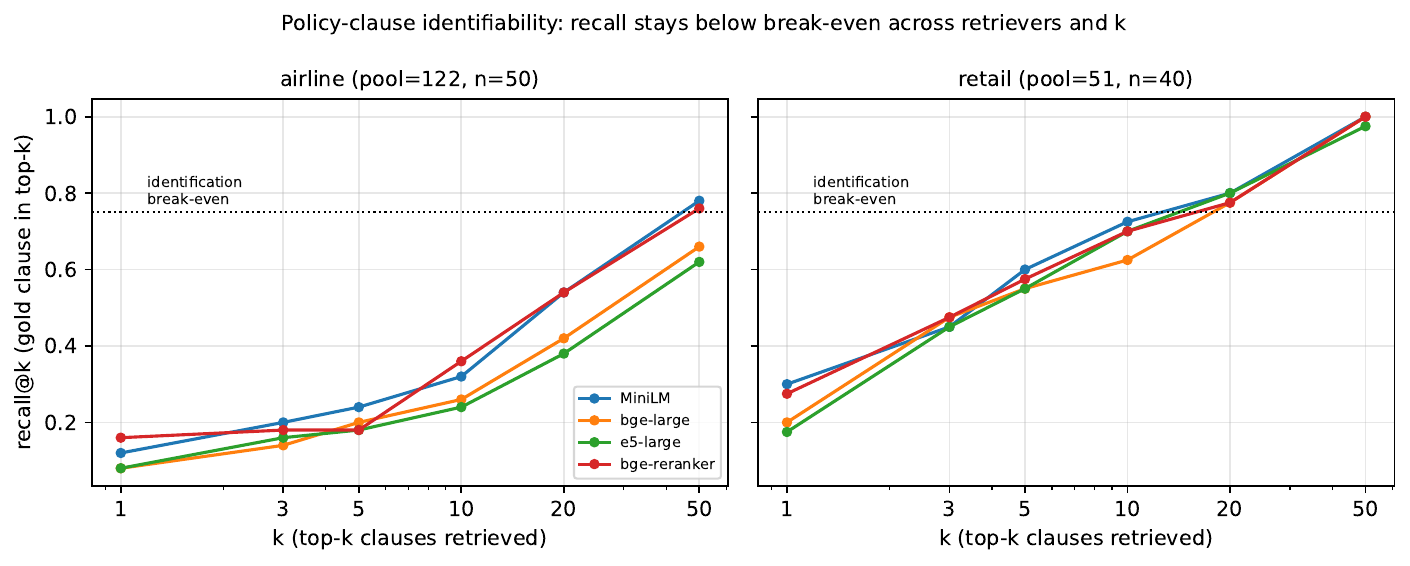}
\caption{\textbf{Policy-clause identifiability profile (recall@$k$).} For each domain, recall@$k$ of the applicable
gold policy clause as a function of $k$ (log scale), for all four off-the-shelf retrievers, under the
trajectory-context query protocol. Dotted line = the $\approx0.75$ diagnostic crossing. On airline (pool $122$)
recall stays far below this crossing until $k\!=\!50$; on retail (pool $51$) it reaches the crossing only near $k\!=\!20$--$50$,
i.e.\ only with large candidate sets comprising a substantial fraction of the policy pool. No retriever,
including the full-pool cross-encoder reranker, identifies the applicable clause at small, ingestable $k$.}
\label{fig:recallcurves}
\end{figure*}

\section{Full representation ablation}
\label{app:repr}
This appendix gives the full representation ablation summarized in Section~\ref{sec:results}. \textbf{Throughout, the
classifier is given the correct applicable policy clause; these are representation effects \emph{given} the rule, not
end-to-end gains} (Section~\ref{sec:identification} replaces this gold-policy condition with retrieved assertions).

\subsection{Primary 3B result}
\begin{figure}[t]
\centering
\includegraphics[width=0.96\columnwidth]{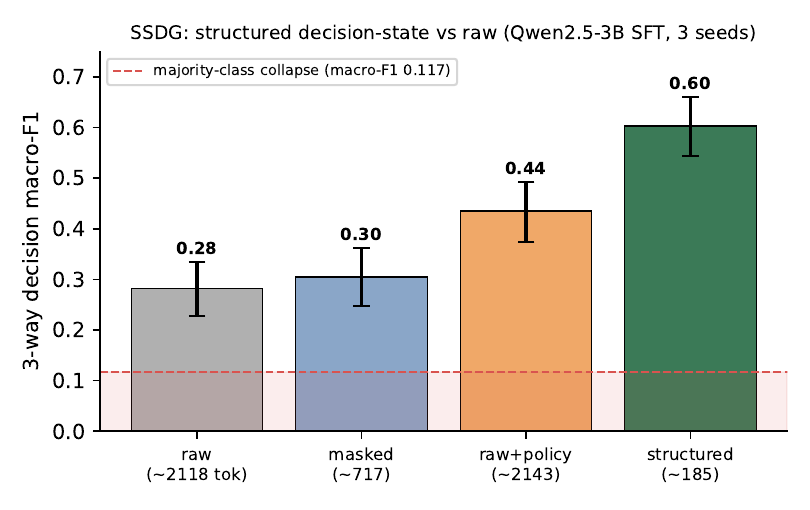}
\caption{Three-way decision macro-F1 by classifier input under the identical SFT recipe (3 seeds, bootstrap CIs). The
dashed line is the majority-class (collapse) macro-F1. Structured decision-state roughly doubles raw.}
\label{fig:main}
\end{figure}

Table~\ref{tab:main} reports the primary \emph{3B} representation result (\S\ref{sec:generalize} shows how it shifts at 7B). Under the
\emph{identical} recipe that produced the refuse-collapse, the structured-state classifier reaches macro-F1 $0.603$ (CI
$[0.543,0.661]$) versus $0.283$ for raw---a paired gap of $+0.320$ (CI $[+0.244,+0.397]$), roughly a doubling.
Crucially, \emph{no} trained representation reproduces the degenerate collapse ($0/12$ runs collapse, vs the
\texttt{ce\_smoke} baseline's refuse-100\%): the collapse was a property of the raw-text small-SFT regime,
and a better representation alone escapes it. Figure~\ref{fig:main} visualizes the ordering.

\begin{table}[t]
\centering\small
\setlength{\tabcolsep}{4pt}
\begin{tabular}{lccc}
\toprule
Gate input (tok) & macro-F1 & 95\% CI & coll. \\
\midrule
\texttt{ce\_smoke}\,$^\dagger$ & $\sim$0.12 & -- & \textbf{3/3} \\
\midrule
raw (2118) & 0.283 & [0.228, 0.334] & 0/3 \\
masked (717) & 0.305 & [0.248, 0.363] & 0/3 \\
raw+policy (2143) & 0.435 & [0.374, 0.491] & 0/3 \\
\textbf{structured (185)} & \textbf{0.603} & [0.543, 0.661] & 0/3 \\
\bottomrule
\end{tabular}
\caption{Main representation ablation (Qwen2.5-3B, LoRA, identical recipe to the committed \texttt{ce\_smoke}
collapse$^\dagger$; 3 seeds; pooled paired bootstrap, 5000 resamples, per-task split). ``tok''=mean input tokens;
``coll.''=seeds that collapsed to one class. Structured decision-state roughly doubles raw macro-F1 (paired
$\Delta{=}{+}0.320$, CI[$+0.244$,$+0.397$]) and \emph{no} trained variant reproduces the refuse-100\% collapse.
$^\dagger$\texttt{ce\_smoke}=the companion postmortem's documented small-data collapse, recomputed from committed
records (raw \texttt{state\_text}, 3 supervision views); macro-F1 $\approx0.12$ at the refuse-100\% collapse
(trained accuracy $0.23$--$0.30$, i.e.\ the majority-class rate).}
\label{tab:main}
\end{table}

\subsection{Controls for brevity, scaffold, label leakage, and output format}
\label{sec:controls}
\begin{table}[t]
\centering\footnotesize
\setlength{\tabcolsep}{4pt}
\begin{tabular}{lc}
\toprule
Control: structured $-$ \emph{X} & $\Delta$ F1 [95\% CI] \\
\midrule
length-matched (raw trunc.) & $+.184$ [$.087,.283$] \\
info-matched (random win.) & $+.180$ [$.027,.330$] \\
structure vs.\ content (shuffled) & $+.313$ [$.169,.449$] \\
\midrule
\multicolumn{2}{l}{\emph{Leakage audit (policy field a leaked label?)}} \\
policy shuffled (wrong policy) & $+.268$ [$.163,.373$] \\
policy generic (contentless) & $+.165$ [$.082,.254$] \\
\bottomrule
\end{tabular}
\caption{Controls (offline frozen-encoder gate). The structured advantage is not brevity (length/info-matched), not
schema scaffold alone (shuffled fields), and not a leaked label: replacing the policy line with a mismatched but
plausible real policy substantially reduces classifier performance. Policy text keyword-matches the gold action only
14--33\%, and the split is task-disjoint. All $\Delta$ are paired-bootstrap macro-F1 gaps (structured minus the named
control); leading zeros are omitted (e.g.\ $.184=0.184$); all CIs exclude zero. Rows are offline-MiniLM-logreg gaps;
the length-matched row is from an earlier $35$-task pilot run and the others from the $195$-task offline ablation, so
the controls are directional rather than from a single jointly fitted model.}
\label{tab:controls}
\end{table}

Table~\ref{tab:controls} collects the frozen-encoder diagnostic controls. The structured advantage survives a
\emph{length-matched} control (raw truncated to the structured length; $+0.184$) and an \emph{information-matched}
control (a random raw window of equal length; $+0.180$), so it is not merely brevity or denoising. It survives
\emph{field-shuffling} (keeping the schema but replacing field contents with another example's; $+0.313$), so it is
not the schema scaffold alone.

\paragraph{Leakage audit.} Because the structured state contains an ``applicable policy'' line, a possible alternative
explanation is that the classifier simply reads a near-label rationale. Three checks argue against this explanation. First, replacing the policy line with
a \emph{mismatched-but-plausible} real policy assertion from another example sharply reduces performance ($\Delta{=}{+}0.268$,
CI $[+0.163,+0.373]$)---the classifier uses the \emph{correct} policy content, not a generic format cue, and a shuffled
real policy is wrong content in the same distribution, so this is not a hidden gold-label leak. Second, the policy
text keyword-matches the gold action only 14--33\% of the time; the classifier must reason policy$+$evidence$\rightarrow$
action. Third, the train/test split is task-disjoint (no test task's policy strings appear in training), so the gain
generalizes to unseen tasks within this domain---though not necessarily to unseen \emph{policy corpora}.

\paragraph{Not output-format compliance.} The raw SFT classifier emits some unparseable continuations on its long inputs.
Re-scoring on the parsed-only subset leaves a large gap (raw parsed-only macro-F1 $0.311$ vs structured $0.603$),
so the advantage is decision quality, not the structured classifier's higher format compliance.

\subsection{How much is policy access vs.\ structured selection?}
We decompose the SFT gain. Appending the elicited policy to raw text (\textbf{raw+policy}) recovers part of the gap
($+0.153$ over raw, CI $[+0.085,+0.220]$), but the structured state still beats raw+policy by $+0.168$
(CI $[+0.101,+0.235]$). So policy access explains roughly half the gain and the compact structured selection of
request/evidence/policy/action-class explains the other half:
\emph{structured decision-state distillation improves over raw trajectories even when policy access is controlled;
the gain is not explained by merely appending policy text. Generic structure alone does not dominate raw}
---indeed, in the frozen-encoder diagnostic probe a \emph{contentless} generic-policy structured state underperforms raw, so the
structured win depends on real elicited content, not formatting in the abstract.

\paragraph{Where the gain concentrates.} On the easy \texttt{allow}-vs-rest contrast, representation barely matters
($0.786$ vs $0.783$); the structured advantage concentrates on the safety-relevant \texttt{refuse}-vs-rest contrast
($0.681$ vs $0.623$), which motivates keeping a distinct \texttt{verify-then-proceed} class rather than a binary
block/allow classifier.

\subsection{Per-class and per-task behaviour}
To check whether aggregate macro-F1 masks class-specific behavior, Table~\ref{tab:perclass} reports
per-class precision/recall/F1 (computed from the SFT predictions, pooled over seeds). Structured improves all three:
\texttt{allow} F1 rises $0.39\to0.63$, \texttt{verify} F1 rises $0.03\to0.55$, and \texttt{refuse} F1 rises
$0.43\to0.64$ while recall reaches $1.00$. Raw+policy improves \texttt{allow} and \texttt{refuse}, but its
\texttt{verify} F1 remains low ($0.17$).
Per task (15 disjoint test tasks, seed 42), the structured-over-raw accuracy delta is positive on \textbf{8} tasks,
tied on \textbf{5}, and negative on \textbf{2}: the gain is broad, not driven by one or two tasks, though the small
per-task counts mean individual tasks are noisy.

\begin{table}[t]
\centering\small
\begin{tabular}{llccc}
\toprule
Input & class & P & R & F1 \\
\midrule
\multirow{3}{*}{raw}
 & allow & 0.50 & 0.32 & 0.39 \\
 & verify & 0.18 & 0.02 & 0.03 \\
 & refuse & 0.31 & 0.70 & 0.43 \\
\midrule
\multirow{3}{*}{raw+policy}
 & allow & 0.82 & 0.39 & 0.53 \\
 & verify & 0.34 & 0.12 & 0.17 \\
 & refuse & 0.46 & 0.89 & 0.60 \\
\midrule
\multirow{3}{*}{\textbf{structured}}
 & allow & 0.59 & 0.67 & 0.63 \\
 & verify & 0.88 & 0.40 & 0.55 \\
 & \textbf{refuse} & 0.47 & 1.00 & 0.64 \\
\bottomrule
\end{tabular}
\caption{Per-class precision/recall/F1 (SFT gate, pooled over 3 seeds; computed from committed predictions). Structured
improves all three classes and reaches \texttt{refuse} recall $=1.00$. Raw has near-zero \texttt{verify} recall, partly
because unparseable outputs count as invalid predictions.}
\label{tab:perclass}
\end{table}

\subsection{Error analysis across decision classes}
\label{sec:erroranalysis}
\begin{table}[t]
\centering\small
\begin{tabular}{lccc}
\toprule
Gold decision & $n$ & raw acc & structured acc \\
\midrule
\texttt{allow} & 24 & 0.33 & \textbf{0.67} \\
\texttt{verify} & 43 & 0.02 & \textbf{0.40} \\
\texttt{refuse} & 18 & 0.61 & \textbf{1.00} \\
\bottomrule
\end{tabular}
\caption{Per-gold-class accuracy (3B SFT, seed 42, committed predictions). In this seed, structured improves over raw
on \emph{all three} classes ($+0.33$ allow, $+0.37$ verify, $+0.39$ refuse), so the aggregate gain is not confined to
one decision category.}
\label{tab:perpolicy}
\end{table}

Breaking accuracy out by the gold decision in seed 42 (Table~\ref{tab:perpolicy}) shows the structured advantage is
not confined to one category: structured improves over raw on \emph{all three} classes ($+0.33$ \texttt{allow},
$+0.37$ \texttt{verify}, $+0.39$ \texttt{refuse}). Inspecting individual cases makes the mechanism concrete. On a
must-refuse task (remove a passenger, which the policy forbids), the raw classifier---given the full $\sim$2{,}000-token
transcript---predicts \texttt{allow}, while the structured classifier, whose input surfaces the line
\emph{``Policy: do not remove passenger\dots''}, correctly predicts \texttt{refuse}. The reverse also occurs: on a
legitimate-proceed task the raw classifier over-refuses while the structured classifier, seeing the satisfied policy condition,
correctly allows. The quantitative gain therefore tracks the proposed mechanism---the classifier performs well when
the decision-relevant policy is explicit in its input, and errs when that signal is buried in raw trajectory text.

\paragraph{Sample size.} The panel contains one decision state from each of $280$ trajectories (test split: $85$ states
from $15$ disjoint tasks). Representation confidence intervals resample decision states, while matched-condition delta
intervals cluster by task. We extract one checkpoint per trajectory to avoid treating correlated checkpoints as
independent observations. Raw remains the weakest input across two model scales and classifier families, and the
structured-over-raw ordering persists under 3B tuning. These comparisons share data and are not independent replications,
but their agreement reduces dependence on any one model or classifier family.
\subsection{Does it hold at 7B and across actors?}
\label{sec:generalize}
We re-run the full ablation at Qwen2.5-7B (Table~\ref{tab:scale}) and test transfer across the trajectory-generating
actor (Table~\ref{tab:crossactor}). This is where the boundary of the effect appears.

\begin{table}[t]
\centering\small
\setlength{\tabcolsep}{4pt}
\begin{tabular}{lccc}
\toprule
 & \multicolumn{2}{c}{shared hyperparameters} & \textbf{tuned} \\
\cmidrule(lr){2-3}\cmidrule(lr){4-4}
Gate input & 3B & 7B & \textbf{3B} \\
\midrule
raw & 0.283 & 0.298 & 0.305 \\
masked & 0.305 & 0.479 & -- \\
\textbf{structured} & \textbf{0.603} & 0.456 & \textbf{0.504} \\
raw+policy & 0.435 & \textbf{0.548} & -- \\
\midrule
structured$-$raw & +0.320 & +0.157 & \textbf{+0.199} \\
\bottomrule
\end{tabular}
\caption{Representation ablation across model scale using shared hyperparameters, plus a 3B sensitivity analysis with
per-cell learning-rate/epoch selection on a held-out task-disjoint validation split (3 seeds). Structured beats raw
at both scales under the shared recipe and after 3B tuning. Only raw and structured were tuned; masked and raw+policy
are shown under the shared recipe for reference.}
\label{tab:scale}
\end{table}

\paragraph{What is robust.} Across \emph{both} scales, raw trajectory text is the worst input ($0.28$--$0.30$
untuned), and getting the applicable policy into the input is the dominant lever: at 7B, simply \emph{appending} the
policy to raw text (\texttt{raw+policy}, $0.548$) is the single best representation, and the leakage controls
(Section~\ref{sec:controls}) show the classifier uses the policy's \emph{content}. Notably, our two methods
\emph{triangulate} the same conclusion from opposite ends: a \emph{frozen-encoder} probe with no fine-tuning
(Table~\ref{tab:controls}, the cross-actor test) and \emph{generative} SFT classifiers (Tables~\ref{tab:main},
\ref{tab:scale}) both rank raw last and reward policy access---so the effect is not an artifact of any single
training setup.

\paragraph{Sensitivity to hyperparameter tuning.} At 3B, we grid-searched learning rate $\times$ epochs (LoRA rank
fixed), selected the best raw and structured configurations on a \emph{held-out validation split} (task-disjoint from
both train and test), and evaluated once on the untouched test set (3 seeds). The structured$>$raw advantage remains
$+0.199$ (Table~\ref{tab:scale}). At 7B, it is $+0.157$ under the shared hyperparameters. We therefore treat tuning as
a 3B sensitivity analysis rather than claiming a two-scale tuned comparison. We report ``best observed
configuration'' rather than a tuned optimum, since the validation split is small ($55$ states / $10$ tasks).

\begin{table}[t]
\centering\small
\begin{tabular}{lc}
\toprule
Gate input & macro-F1 [95\% CI] \\
\midrule
raw & 0.640 [0.49,0.77] \\
masked & 0.818 [0.70,0.92] \\
structured & 0.741 [0.60,0.86] \\
\bottomrule
\end{tabular}
\caption{Cross-actor transfer (offline gate; train on Sonnet-actor decision states, test on Nova-actor, $n{=}50$).
The gate \emph{transfers} across the trajectory-generating model (all inputs $\ge 0.64$), but the structured-vs-raw
advantage is not robust here ($+0.103$, CI $[-0.080,+0.288]$, crosses zero) and masked leads---consistent with the
scale finding that the \emph{specific} structured edge is regime-dependent while abstraction away from raw is more
stable in these checks.}
\label{tab:crossactor}
\end{table}

\paragraph{Cross-actor.} Training on Sonnet-actor decision states and testing on Nova-actor states
(Table~\ref{tab:crossactor}, frozen-encoder probe), the classifier \emph{transfers}---all inputs score $\ge 0.64$ and raw
is again the weakest---but at this small sample ($n{=}50$) the structured-vs-raw margin is not significant
($+0.103$, CI $[-0.080,+0.288]$) and masking leads. The raw-last ordering appears across actors, but the
structured-versus-raw difference is not distinguishable from zero. Which abstraction wins remains underdetermined at
this sample size, consistent with the close masked and structured results at 7B under shared hyperparameters.

\section{Robustness checks for the identification analysis}
\label{app:robust}
We summarise here the supporting checks referenced in Section~\ref{sec:identification}; all use the same
frozen-encoder diagnostic probe.

\paragraph{Gold-injection threshold is robust to the classifier head.} Re-running the gold-injection sweep with a more regularised
logistic head ($C{=}0.1$) and a small MLP ($64$ hidden units) on the same MiniLM features, the structured$-$raw gap
fails to clear zero \emph{even at full gold injection} ($p{=}1.0$): both never robustly beat raw. The balanced-logreg
head ($C{=}1.0$) we report is thus the \emph{most} favourable to the structured arm; none of the tested heads yields a
lower diagnostic crossing.

\paragraph{Representation advantage across encoders.} On the offline probe, the structured$-$raw gap given gold policy
is $+0.115$ (CI $[+0.014,+0.221]$) for MiniLM and $+0.142$ ($[+0.023,+0.260]$) for \texttt{bge-large-en-v1.5}, but only
$+0.038$ ($[-0.092,+0.164]$, not significant) for \texttt{e5-large-v2}. The offline representation effect is therefore
mostly but not universally encoder-robust; our primary representation evidence is the Qwen classifier (Section~\ref{sec:results}),
for which the offline probe is only a low-cost control.

\paragraph{Stronger identification pipelines.} Beyond the four off-the-shelf retrievers, Table~\ref{tab:strongid}
reports three additional retrieval pipelines---query expansion, hybrid first-stage with cross-encoder reranking of the top-20,
and coarse-to-fine hierarchical retrieve. None clears the diagnostic crossing (best recall@5 $0.38$ airline / $0.58$ retail);
the hierarchical variant is the worst, as the coarse section step drops the gold clause. These pipelines test whether
stronger retrieval closes the exact-recall gap; none does in these settings.

\begin{table}[t]
\centering\footnotesize
\setlength{\tabcolsep}{4pt}
\begin{tabular}{@{}l cc cc@{}}
\toprule
& \multicolumn{2}{c}{airline} & \multicolumn{2}{c}{retail} \\
\cmidrule(lr){2-3}\cmidrule(lr){4-5}
Identification pipeline & R@5 & R@10 & R@5 & R@10 \\
\midrule
Best off-the-shelf retriever & 0.24 & 0.36 & 0.60 & 0.72 \\
Query expansion + hybrid & 0.38 & 0.48 & 0.47 & 0.68 \\
Hybrid top-20 $\to$ cross-enc.\ rerank & 0.32 & 0.50 & 0.57 & 0.70 \\
Hierarchical (section $\to$ clause) & 0.22 & 0.26 & 0.35 & 0.47 \\
\midrule
\emph{Break-even (struct.\ beats raw)} & \multicolumn{4}{c}{$\gtrsim 0.75$} \\
\bottomrule
\end{tabular}
\caption{Stronger, additional retrieval pipelines do \emph{not} clear the exact-recall diagnostic. Beyond the four
off-the-shelf retrievers, we test query expansion, a hybrid first-stage with cross-encoder reranking of the top-$20$
candidates, and a coarse-to-fine hierarchical retrieve. The best recall@5 reaches only $0.38$ (airline) / $0.58$
(retail), still well below the $\gtrsim0.75$ diagnostic crossing; hierarchical routing is worse because the coarse step drops
the gold clause. The exact-match shortfall is thus not an artifact of weak/off-the-shelf retrieval.}
\label{tab:strongid}
\end{table}

\paragraph{Per-class identifiability.} Table~\ref{tab:identifclass} breaks airline policy-clause identifiability down by
gold decision class.

\begin{table}[t]
\centering\small
\setlength{\tabcolsep}{5pt}
\begin{tabular}{lcccc}
\toprule
$a^\star$ class & $n$ & recall@5 & recall@10 & gold rank (med) \\
\midrule
allow & 24 & 0.00 & 0.00 & 40/122 \\
verify & 43 & 0.05 & 0.05 & 48/122 \\
\textbf{refuse} & 18 & \textbf{0.50} & \textbf{0.50} & \textbf{7/122} \\
\bottomrule
\end{tabular}
\caption{Policy identifiability by decision class (airline, MiniLM over the $122$-clause pool, request$+$evidence
query). Exact-match recall varies substantially by decision class: recall@5 is $0.00$ for \texttt{allow}, $0.05$ for
\texttt{verify}, and $0.50$ for \texttt{refuse}. The \texttt{refuse} class has the highest exact-match recall,
plausibly because prohibitive clauses use more distinctive language. Per-class $n$ is small, so this analysis is
descriptive. (This breakdown uses the MiniLM retriever; the per-class recalls
therefore aggregate to $\approx0.13$, slightly below the $0.16$ headline figure, which is the stronger hybrid
retriever---the conclusion is unchanged either way.)}
\label{tab:identifclass}
\end{table}

\paragraph{Direct-retrieval seed-level results.} Table~\ref{tab:directpolicyseeds} reports the per-seed macro-F1 values
behind the 3B direct retrieval intervention in Table~\ref{tab:retrievedgate} and the corresponding 7B check.

\begin{table}[t]
\centering\small
\setlength{\tabcolsep}{4pt}
\begin{tabular}{llrrrr}
\toprule
Model & Assertion input & seed 7 & seed 42 & seed 123 & Mean \\
\midrule
3B & Gold governing rule & 0.603 & 0.602 & 0.606 & 0.604 \\
3B & MiniLM top-1 & 0.581 & 0.580 & 0.580 & 0.580 \\
3B & \texttt{bge-large} top-1 & 0.574 & 0.550 & 0.589 & 0.571 \\
3B & Random non-gold & 0.315 & 0.315 & 0.315 & 0.315 \\
3B & No assertion line & 0.206 & 0.206 & 0.206 & 0.206 \\
\midrule
7B & Gold governing rule & 0.456 & 0.478 & 0.467 & 0.467 \\
7B & MiniLM top-1 & 0.431 & 0.447 & 0.472 & 0.450 \\
7B & \texttt{bge-large} top-1 & 0.512 & 0.529 & 0.538 & 0.526 \\
7B & Random non-gold & 0.332 & 0.335 & 0.335 & 0.334 \\
7B & No assertion line & 0.288 & 0.285 & 0.281 & 0.285 \\
\bottomrule
\end{tabular}
\caption{\textbf{Seed-level direct retrieval results.} Each cell is macro-F1 on the same $n{=}85$ states;
means average the three seeds. The 3B aggregate CIs are in Table~\ref{tab:retrievedgate}. At 7B, MiniLM top-1 vs.\
gold is $-0.017$ (task-cluster CI $[-0.201,+0.158]$), while \texttt{bge-large} top-1 vs.\ gold is $+0.059$
(CI $[-0.189,+0.304]$).}
\label{tab:directpolicyseeds}
\end{table}

\paragraph{Observational retrieval--accuracy split.} The retrieval-conditioned top-1 classifier
is not better when the gold clause is in the top-5 ($0.36$, $n{=}11$) than when it is missed ($0.46$, $n{=}74$), so we
treat this correlation as difficulty-confounded and rely on the controlled direct retrieval tests instead.

\subsection{Qualitative categorization of non-matching retrieved assertions}
\label{app:clausecoding}
To characterize the non-gold top-1 retrieved assertions, we label each as \emph{equivalent}, \emph{sufficient},
\emph{partially relevant}, or \emph{inapplicable}. We categorize all $85$ airline retrieval-conditioned states: $79$
non-matches and $6$ exact matches. One annotator saw each assertion pair and the correct decision $a^\star$, but not the
full trajectory. The labels are descriptive; no second annotator reviewed them, and we report no agreement score.
Because the annotator saw $a^\star$, the \emph{sufficient} and \emph{equivalent} counts may be optimistic. Per-item labels,
rationales, and the codebook are committed
(\texttt{experiments/crossenv/nonmatch\_clause\_coding.csv}); the counts below are recomputable from that file.

\begin{table}[h]
\centering
\small
\begin{tabular}{lccccc}
\toprule
Decision & equivalent & sufficient & partially-rel. & inapplicable & total \\
\midrule
\texttt{allow}  & 0 & 0  & 5 & 19 & 24 \\
\texttt{verify} & 2 & 0  & 4 & 37 & 43 \\
\texttt{refuse} & 4 & 10 & 1 & 3  & 18 \\
\midrule
All & 6 & 10 & 10 & 59 & 85 \\
\bottomrule
\end{tabular}
\caption{Single-annotator categories by decision state, not independent task. The $6$ equivalent rows are the exact
matches. Of $79$ non-matches, $59$ are inapplicable, $10$ partially relevant, and $10$ sufficient. The sufficient
states come from two tasks and two unique assertion pairs, both in refuse-to-cancel task families.}
\label{tab:clausecoding}
\end{table}

Two objective, deterministically-recomputable checks accompany the categorization: of the $79$ non-matching
retrievals, $14$ are same-action-class distractors and $65$ are different-class
(\texttt{retrieval\_error\_modes\_results.json}), and the $6$ exact-match rows are exactly the recall@1 hits
($6/85=0.07$).

\paragraph{Interpretation.} Most non-matching top-1 retrievals are not alternate governing rules: $59/79$ are labeled
inapplicable as rules, often because they are booking-fact or state assertions. A decision-useful minority is labeled
\emph{sufficient} ($10$) or \emph{partially relevant} ($10$). This narrows the mechanism described in
Section~\ref{sec:identification}: the retrieved assertion is usually not a substitute \emph{rule}, yet the classifier
shows no detectable macro-F1 loss (Table~\ref{tab:retrievedgate}) and still beats the no-assertion control, consistent
with aggregate context alignment and classifier robustness to some non-rule assertions. A random non-gold assertion reduces performance
($0.315$). The decision-\emph{sufficient} minority concentrates on the safety-relevant \texttt{refuse} class. Here,
exact-match recall underestimates the downstream utility of retrieved benchmark assertions for two reasons: a few
alternate rules are sufficient, and the classifier tolerates context-aligned non-rule text. It does not frequently
recover a decision-equivalent rule.
Because the labels are coarse (3-way), a finer-grained decision task could make retrieval look less tolerable; the
random non-gold control ($0.315$) nonetheless shows label coarseness alone does not produce the no-loss result.

\section{Reproducibility: fine-tuning configuration}
\label{app:repro}
Because several of our main results come from supervised fine-tuning (SFT) of the classifier (Tables~\ref{tab:main},
\ref{tab:perclass}, \ref{tab:scale}), we disclose the training configuration. The frozen-encoder controls
(Tables~\ref{tab:controls}, \ref{tab:crossactor}) use only a balanced logistic-regression head on
\texttt{all-MiniLM-L6-v2} embeddings.

\paragraph{Configuration.} Table~\ref{tab:hp} lists the fine-tuning configuration. Untuned runs share one recipe
($\text{lr}{=}5\!\times\!10^{-5}$, $5$ epochs); tuned 3B runs select lr/epochs on a task-disjoint validation split
(Table~\ref{tab:tunedcfg}).

\begin{table}[t]
\centering\small
\setlength{\tabcolsep}{4pt}
\begin{tabular}{ll}
\toprule
Setting & Value \\
\midrule
\multicolumn{2}{l}{\emph{Fixed across all runs (from trainer source)}} \\
Adapter & LoRA, $r{=}16$, $\alpha{=}32$, dropout $0.05$ \\
LoRA targets & q,k,v,o\_proj \\
Precision & bfloat16 \\
LR warmup & ratio $0.1$ \\
Decoding (eval) & greedy (\texttt{do\_sample{=}False}) \\
Max seq.\ length & $2048$ (state tail-truncated) \\
Grad.\ checkpointing & on \\
Seeds & $\{42,123,7\}$ \\
\midrule
\multicolumn{2}{l}{\emph{Per-model batch (eff.\ batch $4$)}} \\
3B & batch $2\times$accum $2$ \\
7B & batch $1\times$accum $4$ \\
\midrule
\multicolumn{2}{l}{\emph{Untuned baseline runs (E2/E6)}} \\
LR / epochs & $5\!\times\!10^{-5}$ / $5$ \\
Train / test & 195 / 85 (grouped by task) \\
Base model(s) & Qwen2.5-3B / 7B-Instruct \\
\bottomrule
\end{tabular}
\caption{Gate fine-tuning configuration from the training script and committed per-run records.}
\label{tab:hp}
\end{table}

\begin{table}[t]
\centering\small
\begin{tabular}{lccc}
\toprule
Model / repr. & best lr & best ep & test macro-F1 \\
\midrule
3B raw & 5e-5 & 8 & 0.305$\pm$0.013 \\
3B structured & 2e-5 & 3 & 0.504$\pm$0.021 \\
\bottomrule
\end{tabular}
\caption{Tuned 3B configurations. Grid lr$\in\{2,5,10\}\!\times\!10^{-5}$ and
epochs$\in\{3,5,8\}$; best validation cell retrained on fit$+$val and evaluated once on the untouched test ($3$
seeds).}
\label{tab:tunedcfg}
\end{table}

\end{document}